\documentclass[letterpaper]{article} 
\usepackage{aaai25}
\usepackage{times}  
\usepackage{helvet}  
\usepackage{courier}  
\usepackage[hyphens]{url}  
\usepackage{booktabs}
\usepackage{graphicx} 
\usepackage{natbib}  
\usepackage{caption} 
\usepackage{algorithm}
\usepackage{algorithmic}
\usepackage{xcolor}
\usepackage{amssymb}
\usepackage{amsmath}  
\usepackage{amsfonts} 
\usepackage{amssymb}  
\usepackage{cleveref}

\usepackage{newfloat}
\usepackage{listings}
\DeclareCaptionStyle{ruled}{labelfont=normalfont,labelsep=colon,strut=off} 
\floatstyle{ruled}
\newfloat{listing}{tb}{lst}{}
\floatname{listing}{Listing}
\title{Multi-Scale Representation Learning for Image Restoration \\ with State-Space Model}
\author{
    Yuhong He\equalcontrib\textsuperscript{\rm 1}, Long Peng\equalcontrib\textsuperscript{\rm }, Qiaosi Yi\textsuperscript{\rm 3}$^{\dag}$, Chen Wu\textsuperscript{\rm 2}, Lu Wang\textsuperscript{\rm 1}\thanks{Lu Wang and Qiaosi Yi are the corresponding authors.}
}
\affiliations{
    \textsuperscript{\rm 1}Northeastern University, China\\
    \textsuperscript{\rm 2}University of Science and Technology of China\\
    \textsuperscript{\rm 3}The Hong Kong Polytechnic University\\
}

\usepackage{bibentry}

\begin{document}
\maketitle

\begin{abstract}
{Image restoration endeavors to reconstruct a high-quality, detail-rich image from a degraded counterpart, which is a pivotal process in photography and various computer vision systems. In real-world scenarios, different types of degradation can cause the loss of image details at various scales and degrade image contrast. Existing methods predominantly rely on CNN and Transformer to capture multi-scale representations. However, these methods are often limited by the high computational complexity of Transformers and the constrained receptive field of CNN, which hinder them from achieving superior performance and efficiency in image restoration. To address these challenges, we propose a novel Multi-Scale State-Space Model-based (MS-Mamba) for efficient image restoration that enhances the capacity for multi-scale representation learning through our proposed global and regional SSM modules. Additionally, an Adaptive Gradient Block (AGB) and a Residual Fourier Block (RFB) are proposed to improve the network's detail extraction capabilities by capturing gradients in various directions and facilitating learning details in the frequency domain. Extensive experiments on nine public benchmarks across four classic image restoration tasks, image deraining, dehazing, denoising, and low-light enhancement, demonstrate that our proposed method achieves new state-of-the-art performance while maintaining low computational complexity. The source code will be publicly available.}
\end{abstract}

\section{Introduction}
\label{sec:introduction}

{In the process of real-world image transmission, images are inevitably degraded by various factors such as noise, low-light, and adverse weather conditions \cite{AGP, ResFFT}. These types of degradation significantly impair image visibility~\cite{khudjaev2024dformer} and negatively impact downstream vision tasks like autonomous driving \cite{driving} and surveillance \cite{surveilance}. Image restoration, a fundamental task in image processing, aims to reconstruct high-quality and detail-rich images from degraded inputs \cite{mitnet, Chen2018Retinex, CycleISP}. In real-world scenarios, different degradation patterns lead to detail and contrast loss at various scales. For instance, noise disrupts the fine textures of images \cite{deamnet, DAGL}, whereas the different scales of raindrops and rain streaks degrade multi-scale details of images \cite{DualGCN, RCDNet}. Additionally, adverse conditions such as hazy weather and low-light conditions further complicate the extraction of details and contrast \cite{pmnet,yang2021sparse}. Therefore, numerous methods have been proposed to capture contextual information across different scales and model the relationships between them to enhance the details and contrast of the restored images~\cite{MDFEN, maxim, DLINet}.

\begin{figure}[t]
\centering
\begin{minipage}[b]{1.0\linewidth}
  \centering
\vspace{-16pt}
\includegraphics[width=1.0\linewidth]
{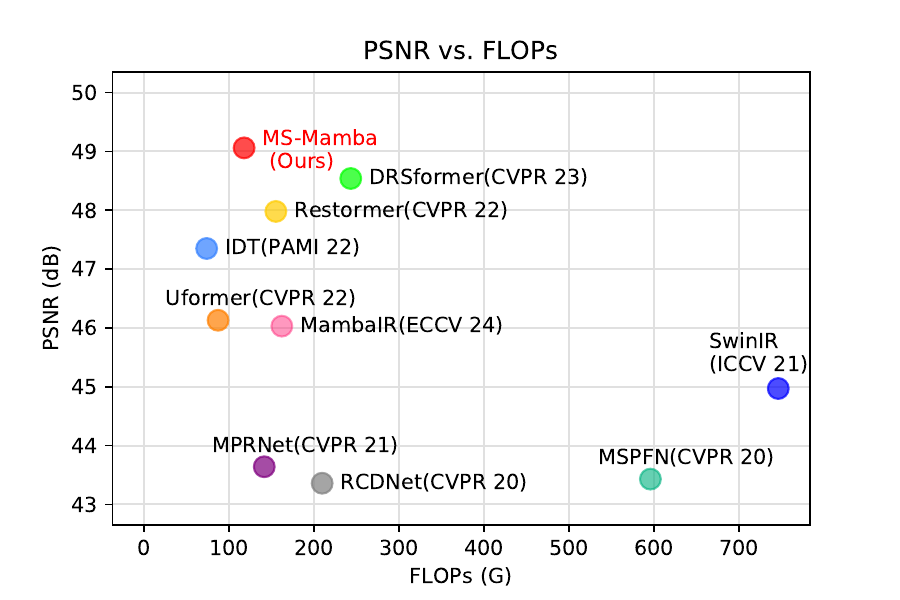}
\end{minipage}
\caption{Model complexity and performance comparison between our MS-Mamba and existing state-of-the-art and classic image restoration methods on the SPA-Data dataset. Our method achieves superior performance while maintaining lower computational costs.}
\label{fig:flops}
\vspace{-18pt}
\end{figure}

{Specifically, previous methods introduce multi-scale convolutional neural network (CNN) modules to enhance the perception of multi-scale information \cite{MSPFN, MPRNet}. However, these modules primarily focus on local feature extraction and struggle to capture global information. To overcome these limitations, researchers propose employing the self-attention mechanism through Transformers to effectively model long-range global features. \cite{dehazeformer, IDT, cai2023retinexformer}. These approaches enable networks to capture global contextual information, providing new perspectives and possibilities for multi-scale representation learning. Recognizing the distinct advantages of Transformers and CNN in modeling long-range dependencies and local features, researchers propose hybrid CNN-Transformer modules to develop multi-scale architecture \cite{DRSformer, LYTNet}. Although hybrid CNN-Transformer methods facilitate networks in handling degradation patterns in complex real-world scenes, they still face the following challenges: a) Transformers have high computational costs and complexity, particularly when processing high-resolution images \cite{liang2021swinir,dehamer}. b) CNN struggles to provide rich local features, which limits its ability to model middle-scale (i.e., regional-level) features \cite{RUAS, MIRNet, VDIR}.

{To address these challenges, we revisit multi-scale representation learning in image restoration and propose a novel Multi-Scale SSM-based (MS-Mamba) representation learning network. It can achieve the best image restoration performance while maintaining low computational cost and high efficiency, as shown in Fig. \ref{fig:flops}. Specifically, we propose a novel Hierarchical Mamba Block~(HMB), which can enhance the network’s capacity for multi-scale representation through the integration of global and regional State-Space Model (SSM) modules and efficient local CNN. In HMB, to efficiently learn global features, we design a global SSM module. This module leverages efficient state-space functions to enhance the network's ability to capture global contextual information. To effectively learn regional-level features, we innovatively propose a regional SSM module in HMB, which integrates multi-scale attention mechanisms within regional windows and significantly improves the network's ability to perceive regional-level features. By leveraging the local extraction capabilities of CNN in HMB, MS-Mamba can capture multi-scale representations. Moreover, to further enhance the capabilities in reconstructing details, a novel Adaptive Gradient Block (AGB) is proposed to explicitly guide the network in modeling detailed features by capturing various directional gradients. Concurrently, a Residual Fourier Block (RFB) is introduced to improve the network's effectiveness in extracting fine textures in the frequency domain. Extensive experiments demonstrate that our MS-Mamba significantly surpasses existing state-of-the-art (SOTA) image restoration methods in four classic image restoration tasks with low computational complexity. }

The contribution can be summarized as follows:
\begin{itemize}
    \item {We propose a novel multi-scale State-Space Model with UNet architecture, MS-Mamba, to effectively extract multi-scale features for high-quality image restoration.}
    \item {We propose a novel Hierarchical Mamba Block, which includes a global and regional State-Space Model (SSM) module to facilitate the capture of global and regional features through state-space functions. Furthermore, an Adaptive Gradient Block (AGB) and a Residual Fourier Block (RFB) are introduced to enhance the network's detail extraction capabilities by capturing gradients in various directions and learning in the frequency domain.}
    \item {Extensive experiments on nine public benchmarks of four image restoration tasks demonstrate our proposed method outperforms existing state-of-the-art methods with low computational complexity.}
\end{itemize}

\section{Related Work}
\subsection{Image Restoration}
Image restoration aims to remove undesired degradations~(\textit{e.g.} noise, blur, haze, rain \textit{etc.}) in corrupted images. With the development of deep learning, lots of works~\cite{DDN, PreNet, SPDNet, 
KinD, peng2021ensemble,peng2024lightweight,he2024latent} have been proposed and gained popularity over traditional methods \cite{Gauss}. In these methods, the encoder-decoder architecture is the most common framework since it can capture multi-scale features. Additionally, other techniques such as residual connections, various attention modules, and dilated convolutions~\cite{RESCAN, peng2020cumulative, 
 mitnet, DRANet, xu2022snr} have been incorporated to extract richer and more important features and expand the receptive field of CNN. Despite their success, CNN-based restoration methods typically have difficulties in effectively modeling global dependencies, leading to performance bottlenecks and insufficient structural reconstruction capabilities. Therefore, transformer-based models have been proposed for image restoration. For instance, Wang et al. introduce a high-performing transformer-based network called Uformer~\cite{uformer} for image restoration. However, it incurs very high computational costs due to the quadratic computational complexity of the self-attention mechanism. To address this issue, window-based transformer networks have been proposed to reduce the computational burden. For example, Liang et al. propose a strong baseline model, SwinIR~\cite{liang2021swinir}, for image restoration based on the Swin Transformer. Similarly, Zamir et al. design Restomer~\cite{zamir2022restormer}, which calculates self-attention alongside the channel dimension. However, these methods cannot completely solve the high computational complexity of the attention mechanism. In this paper, we solve the quadratic complexity problem of the self-attention mechanism through State Space Models.

\subsection{State Space Models}
State Space Models (SSMs), originating from classical control theory, have been introduced to the Natural Language Processing (NLP) and CV (Computer Vision) field as a competitive backbone since SSMs learn global information at a low cost, i.e., with linear complexity. For example, as a landmark work, Mamba~\cite{mamba} used a data-dependent SSM with a selection mechanism to learn global information, which outperforms traditional Transformer-based methods in NLP. With the success of SSMs in NLP tasks, SSMs have also been applied to computer vision tasks~\cite{medical_mamba, liang2024pointmamba, mambair, umamba, zou2024freqmamba}. In image restoration, Guo et al. adopted SSMs to replace the Transformer in SwinIR and proposed MambaIR~\cite{mambair} for image restoration. However, MambaIR only computes SSMs at a single scale, which makes it difficult to effectively capture multi-scale features. Therefore, in this paper, we propose a Hierarchical Mamba Block, which can use the window and global SSM block to perceive regional-level features and global contextual information.

\begin{figure*}[h]
\small
\centering
\begin{minipage}[b]{1.0\linewidth}
  \centering
\includegraphics[width=1.0\linewidth]
{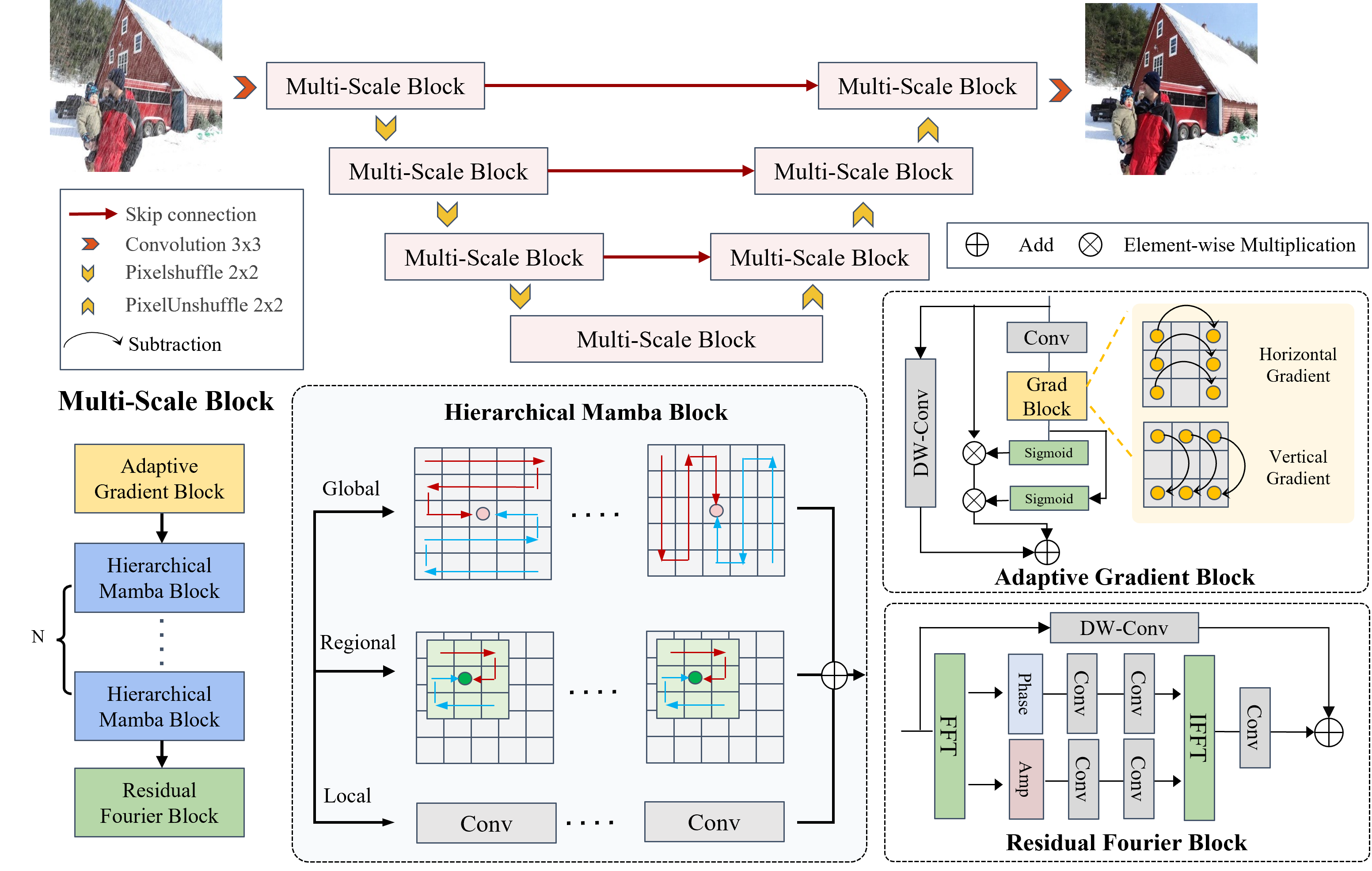}
\end{minipage}
\vspace{-4pt}
\caption{Architecture of our proposed MS-Mamba, which adopts a multi-scale UNet architecture and comprises of the novel Hierarchical Mamba Block, Adaptive Gradient Block, and Residual Fourier Block. }
\label{Mamba}
\vspace{-3pt}
\end{figure*}

\section{Method}
In this section, we present the overall network pipeline of our proposed MS-Mamba, which leverages multi-scale representation learning to handle degradations at various scales in real-world scenarios for restoring image details and contrast. As shown in Fig. \ref{Mamba}, our proposed MS-Mamba employs a multi-scale UNet architecture, which is composed of several fundamental modules dubbed Multi-Scale Blocks. Each Multi-Scale Block primarily contains three key and novel components: the Hierarchical Mamba Block, the Adaptive Gradient Block, and the Residual Fourier Block. These components are proposed to improve the representation capacity of multi-scale details by extracting multi-scale representations and fine-grained gradient information and learning from the frequency domain. Specifically, given input \(I\), we first use a \(3 \times 3\) convolution to extract shallow features \(F_s\), which are then processed by the Multi-Scale Blocks. Finally, a \(3 \times 3\) convolution maps the output features of the multi-scale U-Net to the restored high-quality image \(B\). Next, we will detail the proposed Multi-Scale Block.

\subsection{State Space Models} 
We commence with an exploration of the foundational principles underlying State Space Models (SSM). SSM is conventionally conceptualized as linear time-invariant systems, mapping inputs $x(t) \in \mathbb{R}$ to outputs 
$y(t) \in \mathbb{R}$ via a latent state $h(t) \in \mathbb{R^N}$, where $N$ denotes the dimensionality of the state space. Formally, such systems are characterized by linear ordinary differential equations, as specified in follows:
\setlength\abovedisplayskip{3pt}
\setlength\belowdisplayskip{3pt}
\begin{equation}
\begin{aligned}
h^{\prime}(t)=A h(t)+B x(t) \\
y(t)=C h(t)+D x(t) ,
\end{aligned}
\label{SSM}
\end{equation}
where the parameters include $A \in \mathbb{C}^{N \times N}$, $B,C \in \mathbb{C}^{N}$, and a skip connection $D \in \mathbb{C}^{1}$. Their discrete analogs, exemplified by models like Mamba \cite{mamba}, utilize the zero-order hold (ZOH) method for discretization. This technique enhances the models' capacity to dynamically scan and adjust to input data through a selective scanning mechanism, which is especially advantageous for complex tasks such as image restoration. As a result, MambaIR is proposed. However, it only computes SSMs at a single scale, which limits its ability to capture multi-scale features. Therefore, we propose a Hierarchical Mamba Block to utilize the window and global SSM modules to perceive regional-level features and global contextual information.

\begin{table*}[ht]
\centering
\resizebox{1.0\linewidth}{!}{
\begin{tabular}{c|cccccccccccc}
\toprule
                                 & \multicolumn{2}{c}{Rain200L} & \multicolumn{2}{c}{Rain200H} & \multicolumn{2}{c}{DID-Data} & \multicolumn{2}{c}{DDN-Data} & \multicolumn{2}{c}{SPA-Data} & \multicolumn{2}{c}{Average} \\
Methods                          & PSNR $\uparrow$         & SSIM $\uparrow$          & PSNR $\uparrow$         & SSIM $\uparrow$          & PSNR $\uparrow$         & SSIM $\uparrow$          & PSNR $\uparrow$         & SSIM $\uparrow$          & PSNR $\uparrow$         & SSIM $\uparrow$   & PSNR $\uparrow$         & SSIM $\uparrow$        \\ \midrule
DDN                            & 34.68        & 0.9671        & 26.05        & 0.8056        & 30.97        & 0.9116        & 30.00        & 0.9041        & 36.16        & 0.9457        & 31.57        & 0.9068 \\
RESCAN                         & 36.09        & 0.9697        & 26.75        & 0.8353        & 33.38        & 0.9417        & 31.94        & 0.9345        & 38.11        & 0.9707       & 33.25        & 0.9304 \\
PReNet                        & 37.80         & 0.9814        & 29.04        & 0.8991        & 33.17        & 0.9481        & 32.60        & 0.9459        & 40.16        & 0.9816         & 34.55        & 0.9512\\
MSPFN                          & 38.58        & 0.9827        & 29.36        & 0.9034        & 33.72        & 0.9550         & 32.99        & 0.9333        & 43.43        & 0.9843       & 35.62        & 0.9517 \\
RCDNet                          & 39.17        & 0.9885        & 30.24        & 0.9048        & 34.08        & 0.9532        & 33.04        & 0.9472        & 43.36        & 0.9831        & 35.98        & 0.9554  \\
MPRNet                         & 39.47        & 0.9825        & 30.67        & 0.9110         & 33.99        & 0.9590         & 33.10        & 0.9347        & 43.64        & 0.9844  & 36.17 & 0.9543      \\
SwinIR    & 40.61 & 0.9871 & 31.76 & 0.9151 & 34.07 & 0.9313 & 33.16 & 0.9312 & 44.97 & 0.9890  & 36.91 & 0.9507\\
DualGCN  & 40.73        & 0.9886        & 31.15        & 0.9125        & 34.37        & 0.9620         & 33.01        & 0.9489        & 44.18        & 0.9902    & 36.69 & 0.9604    \\
SPDNet    & 40.50         & 0.9875        & 31.28        & 0.9207        & 34.57        & 0.9560         & 33.15        & 0.9457        & 43.20         & 0.9871   & 36.54 & 0.9594     \\
Uformer & 40.20         & 0.9860         & 30.80         & 0.9105        & 35.02        & 0.9621        & 33.95        & 0.9545        & 46.13        & 0.9913     & 37.22 & 0.9609   \\
Restormer & 40.99        & 0.9890         & 32.00           & 0.9329        & 35.29        & 0.9641        & 34.20         & 0.9571        & 47.98        & 0.9921     & 38.09 & 0.9670   \\
IDT     & 40.74        & 0.9884        & 32.10         & \underline{0.9344}        & 34.89        & 0.9623        & 33.84        & 0.9549        & 47.35        & \underline{0.9930}      & 37.78 & 0.9666   \\
DLINet                         & 40.91        & 0.9886        & 31.47        & 0.9231        & -            & -             & 33.61        & 0.9514        & 44.94        & 0.9885       & 37.73  & 0.9629  \\
DRSformer & \underline{41.23}        & 0.9894        & 32.17        & 0.9326        & \underline{35.35}        & \underline{0.9646}        & \underline{34.35}        & \underline{0.9588}        & \underline{48.54}        & 0.9924   & \underline{38.33} & \underline{0.9676}     \\
MambaIR & 41.13        & \underline{0.9895}        & \underline{32.18}        & 0.9295        & 35.05        & 0.9612        & 34.00        & 0.9554        &   46.03      & 0.9902   & 37.68 &   0.9652  \\
Ours          & \textbf{41.88}           & \textbf{0.9904}                & \textbf{32.24}             & \textbf{0.9365}              & \textbf{35.48}             &  \textbf{0.9656}             & \textbf{34.45}             &  \textbf{0.9601}             & \textbf{49.06}            & \textbf{0.9934} 
& \textbf{38.62}            &\textbf{0.9692} 
\\ \bottomrule            
\end{tabular}
}
\vspace{-3pt}
\caption{Quantitative PSNR $\uparrow$ and SSIM $\uparrow$ comparisons with existing state-of-the-art image deraining methods. The \textbf{bold} and the \underline{underline} represent the best and the second-best performance, respectively. }
\label{Rain} 
\vspace{-3pt}
\end{table*}

\subsection{Hierarchical Mamba Block}
In real-world scenarios, degradations frequently exhibit multi-scale characteristics, requiring effective multi-scale information representation for image restoration. To achieve this, we propose a novel Hierarchical Mamba Block (HMB), which integrates both State Space Models (SSM) and CNN to efficiently capture global, regional, and local representations. While transformer is a common choice for global representation, their computational demands are substantial. Therefore, we employ the Visual State Space (VSS) block \cite{vmamba}, an efficient global SSM that uses a 2D scanning method to model spatial relationships in both horizontal and vertical directions for global sequence modeling.

While the global SSM provides long-range modeling capabilities, it is prone to catastrophic forgetting, leading to insufficient representation of regional and local areas, which are crucial for effective image restoration. To overcome this limitation, we propose a novel regional Window SSM Module, specifically designed to capture regional-scale representations. Specifically, we first partition the image into multiple distinct local windows. For the features within each local window, we perform a linear projection of the sequence and construct sequences in both forward and backward directions to enhance the receptive field. Finally, we compute the output using the SSM.

Despite the aforementioned two SSM modules offer robust global and regional representation modeling capabilities, local representation is essential for perceiving fine-grained degradations. However, directly reducing the window size in the Window SSM Module significantly increases the number of windows, leading to substantial computational complexity. To address this, we employ Convolutional Neural Networks (CNN), a classical approach for extracting local representations, which facilitates the network's ability to learn fine-grained features in local regions. For a given input feature $F_{in}$, the computation process of the HMB can be succinctly represented as:
\begin{equation}
\begin{aligned}
F_{out} = 
\operatorname{GSSM}(\left(F_{in}\right) + \operatorname{RSSM}(\left(F_{in}\right)+ \operatorname{Conv}\left(F_{in}\right) ,
\end{aligned}
\label{gradient}
\end{equation}
where $GSSM$ and $RSSM$ denote the global and regional SSM module, respectively. A comprehensive discussion on the global and regional SSM modules is provided in the \textbf{Appendix A.1}. While the proposed HMB is adept at managing multi-scale features, it lacks the crucial detail modeling capability for image restoration. Therefore, we introduce the Adaptive Gradient Block (AGB) and the Residual Fourier Block (RFB). The AGB is designed to capture gradient details in various directions, whereas the RFB operates in the frequency domain to improve the modeling capacity of multi-scale details.

\begin{figure*}[h]
\centering
\includegraphics[width=1.0\linewidth]{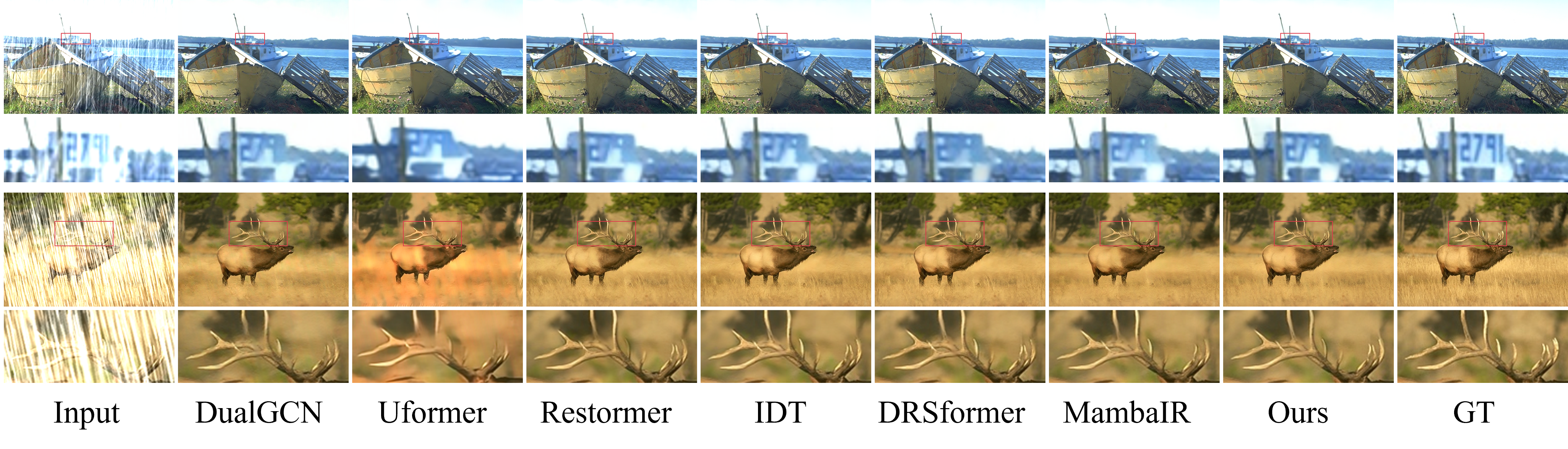}
\centering
\vspace{-3pt}
\caption{Visual comparison on the synthetic rainy images from the Rain200H \cite{JORDER} dataset. Zooming in the figures offers a better view. Our proposed method effectively removes rain streaks and delivers the most visually pleasing results.}
\label{rain200}
\vspace{-3pt}
\end{figure*}

\begin{table*}[ht]
\resizebox{\linewidth}{!}{
\begin{tabular}{c|c|c|c|c|c|c|c|c|c|c}
\toprule
Methods & DeHamer & PMNet &MAXIM &Restormer  &DehazeFormer-L & MDFEN & MITNet      & FocalNet & UMamba & Ours           \\ \midrule
PSNR {\color[HTML]{369DA2} $\uparrow$}    &  36.63 & 38.41 & 38.11 & 38.88  &40.05 & 37.14  & 40.23         & \underline{40.82}    & 40.17  &  \textbf{41.55}              \\
SSIM {\color[HTML]{369DA2} $\uparrow$}    & 0.988 & 0.990 & 0.991 & 0.991 & 0.996 & 0.989  & 0.992         & 0.996    & \underline{0.996}  &   \textbf{0.996}           \\ \bottomrule
\end{tabular}
}
\vspace{-3pt}
\caption{Quantitative comparisons with state-of-the-art dehaze methods on the benchmark SOTS-Indoor \cite{sots}.}
\label{ITS_final_tab} 
\vspace{-3pt}
\end{table*}

\subsection{Adaptive Gradient Block}
To improve the network’s ability to extract fine textures, we propose a novel Adaptive Gradient Block (AGB) to capture gradients in various directions, as shown in Fig.~\ref{Mamba}. Specifically,  a gradient block is introduced in AGB to extract features in both horizontal and vertical directions, as illustrated by the following equations:
\setlength\abovedisplayskip{3pt}
\setlength\belowdisplayskip{3pt}
\begin{equation}
\begin{aligned}
F_{x} &= F_{in}(x+1, y) - F_{in}(x-1, y), \\
F_{y} &= F_{in}(x, y+1) - F_{in}(x, y-1), \\
\nabla F &= \left(F_{x}, F_{y}\right), \\
G &= \|\nabla F\|_{2} ,
\end{aligned}
\label{gradient}
\end{equation}
where $F_{in}$ denotes the input feature, and \(G\) represents the gradient map of $F_{in}$. The variables \(x\) and \(y\) refer to the coordinate indices. Then, leveraging the gradient information from different directions, the AGB generates channel and spatial detail attention maps through the sigmoid activation function to provide the network with the ability to perceive spatial details and detail-related channels, thereby improving the network's overall detail reconstruction capability for degraded images. The above process can be represented as:

\setlength\abovedisplayskip{3pt}
\setlength\belowdisplayskip{3pt}
\begin{equation}
\begin{array}{l}
\begin{aligned}
F_{in} & = \text{ReLU}(\text{Conv}(G)), \\
\mathbf{A}_{c} & = f_{ca}(F_{in}) \in \mathbb{R}^{C \times 1 \times 1}, \\
\mathbf{A}_{s} & = f_{sa}(F_{in}) \in \mathbb{R}^{1 \times H \times W}, \\
\hat{F} & = \mathbf{A}_{s}(\mathbf{A}_{c}(I)) + \text{DW-Conv}(I) ,
\end{aligned}
\end{array}  
\label{adaptive gradient}
\end{equation}
where $\mathbf{A}_{c}$, $\mathbf{A}_{s}$ are the channel and spatial attention maps, respectively. DW-Conv denotes the depth-wise convolution.

\subsection{Residual Fourier Block}
Recognizing that the spatial domain predominantly resides in low-frequency information, rendering high-frequency details challenging for network capture, we propose a novel Residual Fourier Block (RFB) to enhance fine texture extraction in the frequency domain, as shown in the bottom right corner of Fig. \ref{Mamba}. Specifically, the RFB transforms the input feature into the Fourier domain, yielding separate amplitude and phase components. Recognizing the distinct contributions of the amplitude and phase components in degradation, they are processed separately in RFB. Two sequential \(1 \times 1\) convolutional layers, followed by the ReLU activation function, are used for the processing. After that, the processed amplitude and phase components are transformed back into the spatial domain, which is then followed by a \(3 \times 3\) convolutional layer. Additionally, a residual depth-wise convolution consolidates frequency information, aiding in the preservation of high-frequency details crucial for image clarity often diminished during restoration.

\subsection{Loss Function}

Following previous works \cite{zamir2022restormer, DRSformer}, we utilize the common L1 loss \(\mathcal{L}_{1}\) for training. Furthermore, to enhance the network's ability to capture details, we also employ the edge loss \(\mathcal{L}_{edge}\) and the frequency loss \(\mathcal{L}_{fft}\). The total loss is presented as follows:

\begin{equation}
\label{loss}
\begin{aligned}
\mathcal{L}_{total} &= \lambda_{1} \mathcal{L}_{1}(\mathcal{B}, \mathcal{B}_{gt}) + \lambda_{2} \mathcal{L}_{edge}(\mathcal{B}, \mathcal{B}_{gt}) \\
&+ \lambda_{3} \mathcal{L}_{fft}(\mathcal{B}, \mathcal{B}_{gt}) ,
\end{aligned}
\end{equation}
where $\mathcal{B}$ and $\mathcal{B}_{gt}$ denote the predicted output and the corresponding ground truth, respectively. The parameters $\lambda_{1}$, $\lambda_{2}$, and $\lambda_{3}$ are balancing factors. In our experiments, we set $\lambda_{1}$, $\lambda_{2}$, and $\lambda_{3}$ to 1, 0.1, and 0.05, respectively.

\begin{table}[h]
\resizebox{\linewidth}{!}{
\begin{tabular}{c|cccc}
\toprule
                                 & \multicolumn{2}{c}{LOL-v2-real } & \multicolumn{2}{c}{LOL-v2-syn} \\
Method                           & PSNR {\color[HTML]{369DA2} $\uparrow$}      & SSIM {\color[HTML]{369DA2} $\uparrow$}    & PSNR {\color[HTML]{369DA2} $\uparrow$}      & SSIM {\color[HTML]{369DA2} $\uparrow$}     \\ \midrule
RetinexNet               & 15.47       & 0.567         & 17.13         & 0.798          \\
KinD                     & 14.74       & 0.641         & 13.29         & 0.578          \\
RUAS                     & 18.37       & 0.723         & 16.55         & 0.652          \\
Uformer                  & 18.82       & 0.771        & 19.66         & 0.871          \\
Sparse                   & 20.06        & 0.816         & 22.05         & 0.905          \\
Restormer                & 19.94       & 0.827        & 21.41         & 0.830           \\
MIRNet                   & 20.02       & 0.820         & 21.94         & 0.876          \\
SNR-Net                  & 21.48       & \underline{0.849}        & 24.14         & 0.928          \\
Retinexformer            & 22.80       & 0.840        & \underline{25.67}         & 0.930           \\
FreqMamba                & -       & -       & 24.46         & \underline{0.936}    \\ 
LYT-Net                  & \underline{22.93}       & 0.840        & 23.33         & 0.905          \\ \midrule
Ours & \textbf{26.69} & \textbf{0.896} & \textbf{26.62} & \textbf{0.947}      
       \\ 
 \bottomrule 
\end{tabular}
}
\vspace{-3pt}
\caption{Comparisons of low-light image enhancement methods on the LOL-v2 \cite{LOLv2} benchmarks.}
\label{LOL_tab} 
\vspace{-3pt}
\end{table}

\begin{figure*}[h]
\centering
\includegraphics[width=1.0\linewidth]{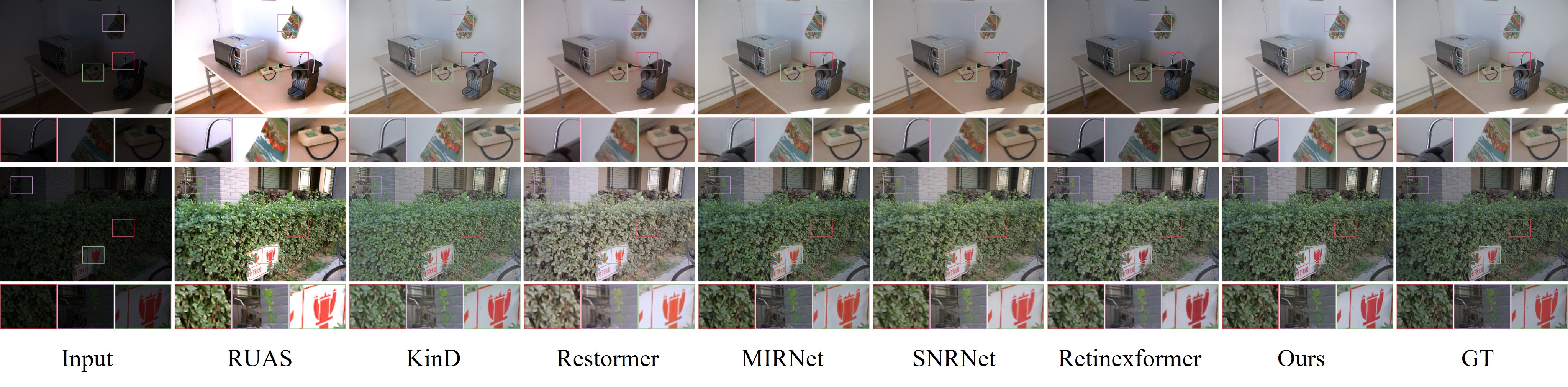}
\centering
\vspace{-3pt}
\caption{Visual comparison on the real low-light images from the LOLv2~\cite{LOLv2} dataset. Zooming in the figures offers a better view. Our proposed method delivers the most visually pleasing results.}
\label{LOLv2_real_fig}
\vspace{-3pt}
\end{figure*}

\begin{figure}[h]
\centering
\includegraphics[width=1.0\linewidth]{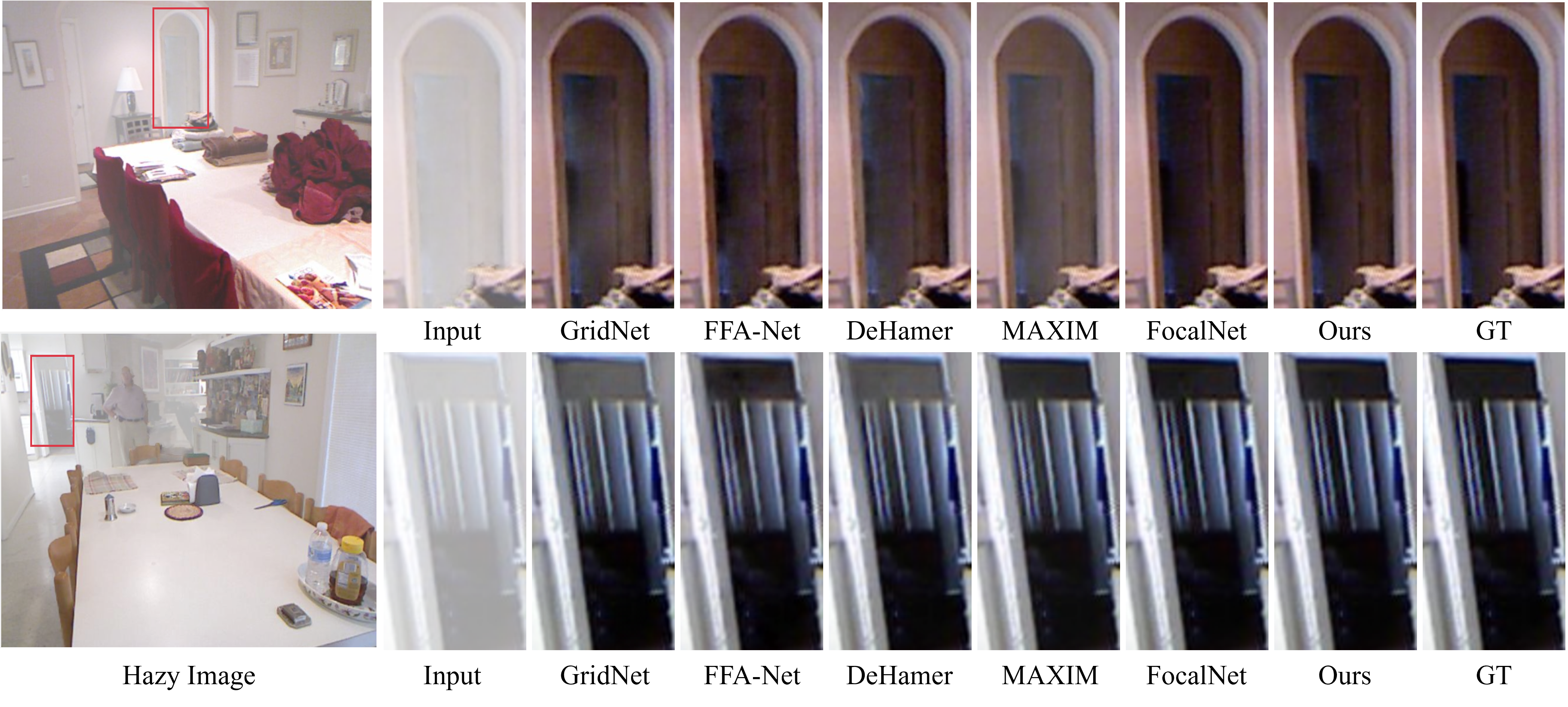}
\centering
\vspace{-3pt}
\caption{Image dehasing visual comparison on the SOTS-indoor \cite{sots} datasets. Best viewed on screen.}
\label{ITS_fig}
\vspace{-3pt}
\end{figure}

\begin{table*}[h]
\resizebox{\linewidth}{!}
{
\begin{tabular}{c|c|c|c|c|c|c|c|c|c|c}
\toprule
Methods & CycleISP & MIRNet &DeamNet & MPRNet  & DAGL & Uformer & VDIR      & DRANet & MambaIR & Ours           \\ \midrule
PSNR {\color[HTML]{369DA2} $\uparrow$}   & 39.52 & 39.72 & 39.47 & 39.71  & 39.77  & 39.77 & 39.26  & 39.50         & \underline{39.89}   &  \textbf{39.92}              \\
SSIM {\color[HTML]{369DA2} $\uparrow$}   &0.957 & 0.959 & 0.957 & 0.958 & 0.959 & 0.959  & 0.955         & 0.957    & \underline{0.960}  &   \textbf{0.960}           \\ \bottomrule
\end{tabular}
}
\vspace{-3pt}
\caption{Quantitative comparisons of different denoising models on the SIDD \cite{sidd} benchmark. }
\label{SIDD_tab} 
\vspace{-3pt}
\end{table*}

\begin{table*}[h]
\centering
\resizebox{\linewidth}{!}
{
\begin{tabular}{c|cccccccccc}
\toprule


Methods & DeHamer   & Uformer  & Restormer & MIRNet & IDT    & DehazeFormer-L & DRSformer & UMamba & MambaIR & Ours      \\ \midrule
Params(M) {\color[HTML]{369DA2} $\downarrow$}  & 132.450    & 50.880   & 26.127 & 31.760 & 16.420   & 25.452          & 33.660     & 19.250  & 31.506    &  \textbf{19.640}     \\
FLOPs(G) {\color[HTML]{369DA2} $\downarrow$}  & 48.926    & 86.891  & 154.882 & 785.000 & 73.474  & 279.624        & 242.989   & 173.550 &  161.902 &  \textbf{117.524}       \\ \bottomrule
\end{tabular}
}
\vspace{-3pt}
\caption{Comparisons of model complexity against state-of-the-art methods. The input size is $256 \times 256$ pixels. }
\label{flops} 
\vspace{-3pt}
\end{table*}

\begin{table}[h]
\centering
\begin{tabular}{ccc|cc}
\toprule
Global & Regional & Local & PSNR {\color[HTML]{369DA2} $\uparrow$}  & SSIM {\color[HTML]{369DA2} $\uparrow$}    \\ \midrule
$\checkmark$                                      &          &       & 40.85 & 0.9844 \\
                                        & $\checkmark$       &       & 41.09 & 0.9888 \\
                                        &          & $\checkmark$    & 36.76 & 0.9757 \\
$\checkmark$       & $\checkmark$ &  & 41.75 & 0.9901\\
$\checkmark$       &  & $\checkmark$ & 41.41 & 0.9898\\
       & $\checkmark$ & $\checkmark$ & 41.55 & 0.9900\\
$\checkmark$                                      & 

$\checkmark$       & $\checkmark$    & \textbf{41.88} & \textbf{0.9904} \\ \bottomrule
\end{tabular}
\vspace{-3pt}
\caption{Ablation study for Hierarchical Mamba Block. }
\vspace{-3pt}
\label{HMB} 
\end{table}

\section{Experiments}
\subsection{Experimental Settings}

{\textbf{Datasets.} To comprehensively verify the superiority of our method in image restoration, we conduct comparison experiments on nine public benchmarks across four classic image restoration tasks. For Image Deraining, following \cite{DRSformer}, we evaluate on five public benchmarks: Rain200L/H, DID-Data, DDN-Data, and SPA-Data, where SPA-Data is a real-world dataset. For Image Dehazing, following \cite{dehazeformer}, we evaluate on the public benchmark SOTS-indoor. For Low-light Image Enhancement, we evaluate on the public LOL-v2-real and LOL-v2-syn benchmarks, following \cite{cai2023retinexformer}. For Image Denoising, following \cite{mambair}, we evaluate on the public real-world SIDD benchmark. More details on the training and test sets are presented in the \textbf{Appendix A.3}.}

{\textbf{Evaluation Metrics.} We use PSNR and SSIM to evaluate the performance. Specifically, we follow \cite{zamir2022restormer} to evaluate on the Y channel for image deraining and denoising. For image dehazing and low-light enhancement, we evaluate on the RGB channel, following \cite{focal} and \cite{cai2023retinexformer}, respectively. Note that higher values of PSNR and SSIM indicate better restoration quality.}

{\textbf{Implementation Details.} Our model employs a 4-level encoder-decoder U-Net architecture, with the number of blocks in each stage set to [2, 2, 2, 2] and the number of feature channels \(C\) set to 48. The window size for the regional SSM Module is configured as [16, 16, 8, 8]. Network parameters are optimized using the AdamW optimizer, starting with an initial learning rate of  \(3 \times 10^{-4}\) for the first 92K iterations,  followed by a reduction to \(1 \times 10^{-6}\) over the subsequent 208,000 iterations, employing a cosine annealing strategy. Additionally, random flips and rotations are applied for data augmentation. Experiments are conducted using the PyTorch framework on four NVIDIA GeForce RTX 3090 GPUs. More details are available in the \textbf{Appendix A.2}.}

{\textbf{Comparisons with State-of-the-art Methods.} \textbf{\textit{Image Deraining}}: we compare our method against fifteen state-of-the-art methods, including: DDN, RESCAN, PReNet, MSPFN, RCDNet, MPRNet, SwinIR, DualGCN, SPDNet, Uformer, Restormer, IDT, DLINet, DRSformer and MambaIR. \textbf{\textit{Image Dehazing}}: we conduct comparisons with nine state-of-the-art methods, including: DeHamer, PMNet, MAXIM, Restormer, DehazeFormer, MDFEN, MITNet, FocalNet and UMamba.
\textbf{\textit{Low-light Enhancement}}: we evaluate our approach against eleven state-of-the-art methods, including: Uformer, RetinexNet, Sparse, RUAS, KinD, Restormer, MIRNet, SNRNet, Retinexformer, FreqMamba and LYTNet.
\textbf{\textit{Image Denoising}}: we evaluate our approach against nine state-of-the-art methods, including: CycleISP, MIRNet, DeamNet, MPRNet, DAGL, Uformer, VDIR, DRANet and MambaIR.} 

\subsection{Quantitative and
Qualitative Results}

To validate the effectiveness of our proposed method, we conduct experiments on four restoration tasks across a total of nine public benchmarks, as shown in Tables \ref{Rain}, \ref{ITS_final_tab}, \ref{LOL_tab}, \ref{SIDD_tab}.
\textbf{\textit{Image Deraining:}} Our method achieves the best performance on all five benchmarks for image deraining. For example, on Rain200L, our method surpasses the existing state-of-the-art (SOTA) method DRSformer by 0.65 dB in PSNR. For the real-world dataset SPA-Data, our MS-Mamba outperforms existing methods by 0.52 dB. \textbf{\textit{Image Dehazing:}} Our method also achieves the best performance on the SOTS-Indoor benchmarks for image dehazing. For example, MS-Mamba demonstrates a 1.38 dB improvement over the SSM-based method Umamba. \textbf{\textit{Low-light Image Enhancement:}} As indicated in Table \ref{LOL_tab}, our model consistently achieves the best performance across all variants of the LOLv2 dataset. For instance, MS-Mamba achieves a 2.16 dB gain over the SSM-based method FreqMamba on the LOL-v2-syn benchmark. \textbf{\textit{Image Denoising:}} Similarly, our method achieves the best performance on the real-world SIDD benchmark for image denoising. Additionally, to demonstrate the visual superiority of the proposed method, we present some visual comparisons on the Rain200H deraining dataset in Fig. \ref{rain200}, the SOTS-indoor dehazing dataset in Fig. \ref{ITS_fig}, and the low-light enhancement LOLv2 dataset in Fig. \ref{LOLv2_real_fig}. These results demonstrate the superiority and generality of the proposed method across various image restoration tasks. More visualization comparisons are presented in the \textbf{Appendix A.5}.

\subsection{Comparison of model complexity} 
{In practical deployment, model complexity is of paramount importance. Therefore, we conduct a comparative analysis of our method against existing state-of-the-art approaches in terms of model parameters and floating point operations (FLOPs), as presented in Table. \ref{flops}. The results indicate that, compared to all existing methods, our model exhibits fewer parameters and lower FLOPs while achieving superior performance. For example, compared to the SSM-based method MambaIR, our model reduces the number of parameters by 11.866 M and has fewer FLOPs. Moreover, our FLOPs are reduced by 56.026 G compared to Umamba. This highlights the efficiency and practicality of our proposed method for real-world applications.}

\subsection{Ablation Study}
To validate the effectiveness of our proposed Hierarchical Mamba Block, we conduct ablation experiments on the Rain200L dataset. Specifically, we remove each branch of the block and consider combinations of any two branches, subsequently evaluating their performance. The results are presented in Table \ref{HMB}. It can be observed that each scale is crucial for the HMB. For instance, when using only the global and regional scales, the PSNR drops by 1.03 dB and 0.79 dB, respectively; when using only the local scale, the PSNR decreases significantly by 5.12 dB. Moreover, when any two of the three scales are present, the performance also declines. These experimental results demonstrate the importance of multi-scale information in the proposed HMB. Additional ablation studies, including various AGB variants, the effectiveness of RFB, and the loss function configurations, are detailed in \textbf{Appendix A.4}.

\subsection{User Study}
To validate the superiority of our method on real-world scenes, a user study is conducted. Specifically, we randomly select 10 rainy images from the Real127 dataset \cite{DIDMDN}. Ten participants rate each image from 0 (not removed at all) to 10 (very clean). The aggregated results, as shown in Fig. \ref{fig:user_study}, indicate that existing methods struggle with adaptive rain removal, resulting in lower user satisfaction. In contrast, our method effectively eliminates degradations, achieving the highest average score of 7.38, and demonstrating superior generalization in real-world scenarios.

\begin{figure}[t]
\centering
\includegraphics[width=1.0\linewidth]{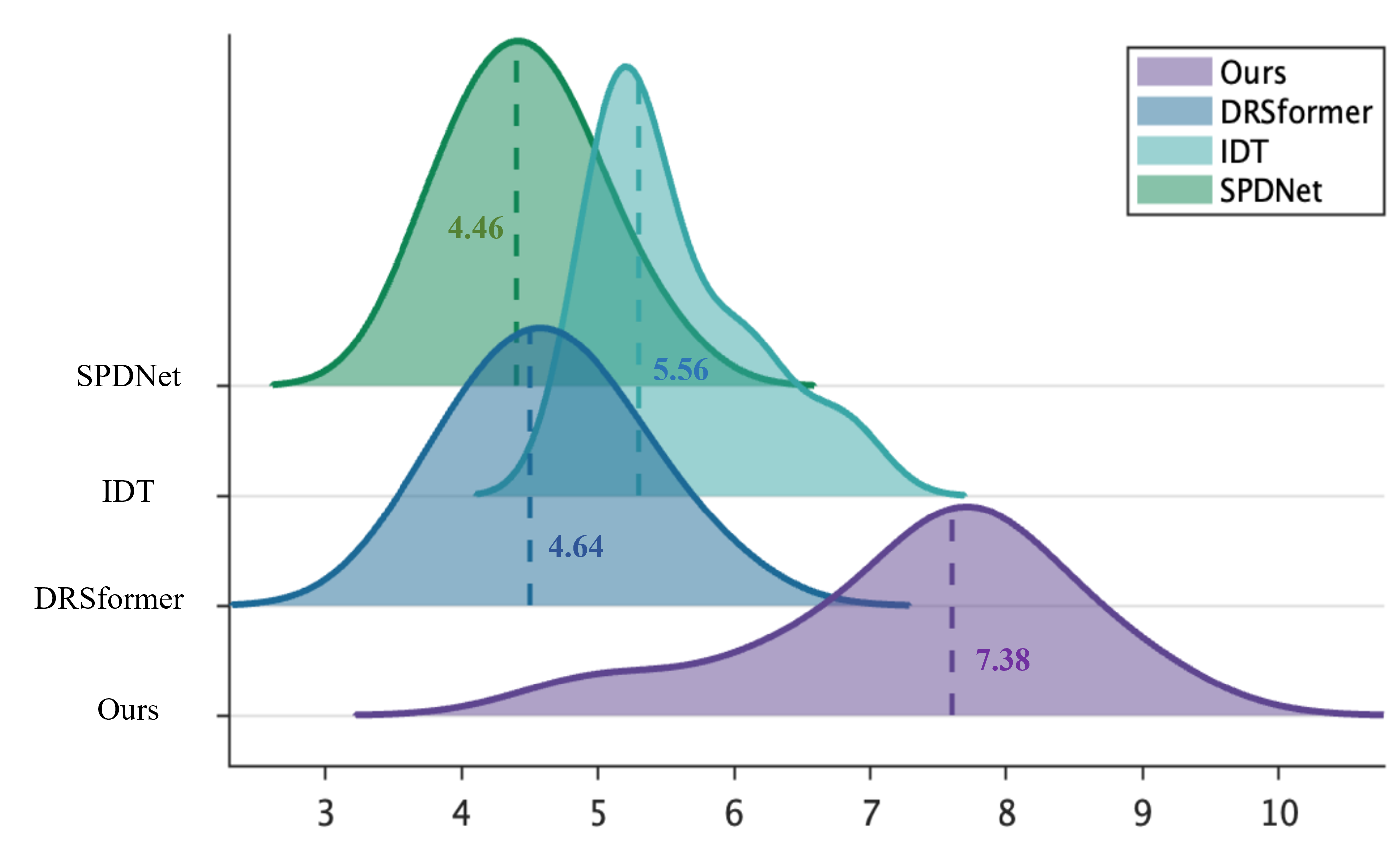}
\centering
\vspace{-3pt}
\caption{Results of the user study on the Real127 Dataset. }
\label{fig:user_study}
\vspace{-3pt}
\end{figure}

\section{Conclusion}
{In this paper, we propose MS-Mamba, an efficient multi-scale SSM with U-Net architecture designed to effectively capture multi-scale features for high-quality image restoration. MS-Mamba effectively captures global contextual information through the proposed global SSM module, enhances regional feature perception capacity via the proposed regional SSM module, and extracts local features using CNN. Additionally, an Adaptive Gradient Block (AGB) and a Residual Fourier Block (RFB) are introduced to enhance the network's detail extraction capabilities by capturing gradients from various directions and learning in the frequency domain. Extensive experiments demonstrate that MS-Mamba significantly outperforms existing SOTA methods on nine public benchmarks and real scenarios across four classic image restoration tasks while maintaining low computational complexity.}

\bibliography{aaai25}
\end{document}